\pdfoutput=1

\documentclass[11pt]{article}

\usepackage[final]{acl}

\usepackage{times}
\usepackage{latexsym}

\usepackage[T1]{fontenc}

\usepackage[utf8]{inputenc}

\usepackage{microtype}

\usepackage{inconsolata}

\usepackage{graphicx}

\usepackage{latexsym}
\usepackage[dvipsnames]{xcolor}
\usepackage{xcolor}
\usepackage{pxrubrica}        \usepackage{url}
\usepackage{tabularx}
\usepackage{tabularray}
\usepackage{multirow}
\usepackage{array}
\usepackage{here}
\usepackage{tabularray}
\usepackage{tablefootnote}
\usepackage{CJKutf8}
\usepackage{mathrsfs}
\usepackage{euscript}

\usepackage{footmisc}
\usepackage{bm}
\usepackage{float}
\usepackage{multicol}
\usepackage{arydshln}
\usepackage{enumitem}
\usepackage{amssymb}
\usepackage{amsmath}

\newcommand{\comment}[1]{\textcolor{black}{#1}}
\newcommand{\crcomment}[1]{\textcolor{black}{#1}}
\newcommand{\mini}[1]{{  \fontsize{8pt}{8pt}  \selectfont            #1                   }}

\newcommand{\tabH}{\rule{0pt}{2.1ex}}
\newcommand{\bhline}{\noalign{\hrule height 1.2pt}}

\newcommand{\footlink}[1]{\footnote{\url{#1}}}

\setlist[description]{
  topsep=5pt,            }
\setlist[enumerate]{
  topsep=1pt,          itemsep=0.5pt,       }
\usepackage{titlesec}

\titlespacing*{\paragraph}{0pt}{0.5\baselineskip}{0.5\baselineskip}

\title{FrameEOL:\\Semantic Frame Induction using Causal Language Models}

\author{
 \textbf{Chihiro Yano\textsuperscript{1}},
 \hspace{4ex}
 \textbf{Kosuke Yamada\textsuperscript{1,2}},
 \hspace{4ex}
 \textbf{Hayato Tsukagoshi\textsuperscript{1}},
 \\
 \textbf{Ryohei Sasano\textsuperscript{1}},
 \hspace{4ex}
 \textbf{Koichi Takeda\textsuperscript{3}}
\\
 \textsuperscript{1}Nagoya University
 \hspace{2ex}
 \textsuperscript{2}CyberAgent
   \hspace{2ex}
 \textsuperscript{3}National Institute of Informatics
\\
 \texttt{yano.chihiro.j3@s.mail.nagoya-u.ac.jp},
 \hspace{4ex}
 \texttt{yamada\_kosuke@cyberagent.co.jp}\\
 \texttt{tsukagoshi.hayato.r2@s.mail.nagoya-u.ac.jp}\\
 \texttt{sasano@i.nagoya-u.ac.jp}
 \hspace{4ex}
 \texttt{takedasu@nii.ac.jp}
}

\begin{document}
\maketitle
\begin{abstract}
Semantic frame induction is the task of clustering frame-evoking words according to the semantic frames they evoke.
\comment{In recent years, leveraging embeddings of frame-evoking words that are obtained using masked language models (MLMs) such as BERT has led to high-performance semantic frame induction.}
\comment{Although causal language models (CLMs) such as the GPT and Llama series succeed in a wide range of language comprehension tasks and can engage in dialogue as if they understood frames, they have not yet been applied to semantic frame induction.}
We propose a new method for \comment{semantic frame induction} based on CLMs. 
Specifically, we introduce FrameEOL, a prompt-based method for obtaining \textbf{F}rame \textbf{E}mbeddings that outputs \textbf{O}ne frame-name as a \textbf{L}abel representing the given situation.\footnote{FrameEOL is a method inspired by PromptEOL, and its name includes the string “EOL” to reflect this connection. However, it should be noted that what “EOL” stands for differs between the two methods.}
To obtain embeddings more suitable for frame induction, we leverage in-context learning (ICL) and deep metric learning (DML).
Frame induction is then performed by clustering the resulting embeddings.
\comment{
Experimental results on the English and Japanese FrameNet datasets demonstrate that the proposed methods outperform existing frame induction methods.
\crcomment{
In particular, for Japanese, which lacks extensive frame resources, the CLM-based method using only 5 ICL examples achieved comparable performance to the MLM-based method fine-tuned with DML.
}
}

\end{abstract}

\section{Introduction}
Frame semantics~\cite{Frame_Semantics} considers access to world knowledge that is related to a word to be essential for understanding the word.
Consider the following sentences:

\begin{enumerate}
    \item He \textit{lost} the gold medal by just .02 points. \label{EX::Finish_competition}
    \item He \textit{lost} his gold medal at the restaurant. \label{EX::Losing}
\end{enumerate}

\noindent Both sentences contain the same verb \textit{lost}.
However, Sentence (\ref{EX::Finish_competition}) conjures an image of ``\textsl{a competition comes to an end, with someone losing against an opponent according to their score},'' while Sentence (\ref{EX::Losing}) evokes a situation where ``\textsl{someone may have unintentionally misplaced a possession}.''
When reading such sentences, humans recognize the different meanings of \textit{lost} from the context and understand each sentence while supplying the information necessary to understand the situation, such as \textsc{competition} or \textsc{opponent} for Sentence~(\ref{EX::Finish_competition}).
In frame semantics, such background knowledge is referred to as a semantic frame and words like \textit{lost} are called frame-evoking words.
The verb \textit{lost} in Sentence (\ref{EX::Finish_competition}) corresponds to the \textsc{Finish\_Competition} frame, which involves a competition coming to an end with an outcome determined by a score or ranking.
In contrast, the verb \textit{lost} in Sentence (\ref{EX::Losing}) corresponds to the \textsc{Losing} frame, which describes a situation where someone loses his or her possession.

Semantic frame knowledge is being developed for many languages.
The representative resource is FrameNet~\cite{FrameNet}, a large-scale semantic frame resource that was manually developed for English. 
In addition, resource development following FrameNet is in progress for several languages, including Brazilian Portuguese, Chinese, Hebrew, Japanese, Korean, and Swedish~\cite{mFrameNet}.
However, manual resource development entails enormous costs, making it challenging to construct such large-scale frame knowledge.
Accordingly, automatic construction of frame knowledge from massive text corpora is an active research area. 
Representative of such studies are those on frame induction: the task of clustering a set of words given together with their context according to the semantic frames they evoke. 
In recent years, several methods \cite{semeval_BERT,semeval2019-2_1st,YAMADA_mask_2} that leverage masked language models (MLMs) such as BERT~\cite{BERT} to obtain frame embeddings have been proposed. 
Frame embeddings are vector representations that capture the semantic frame evoked by a given word and are used to induce semantic frames.

\begin{figure}[t]
    \centering
    \includegraphics[width=1.0\linewidth]{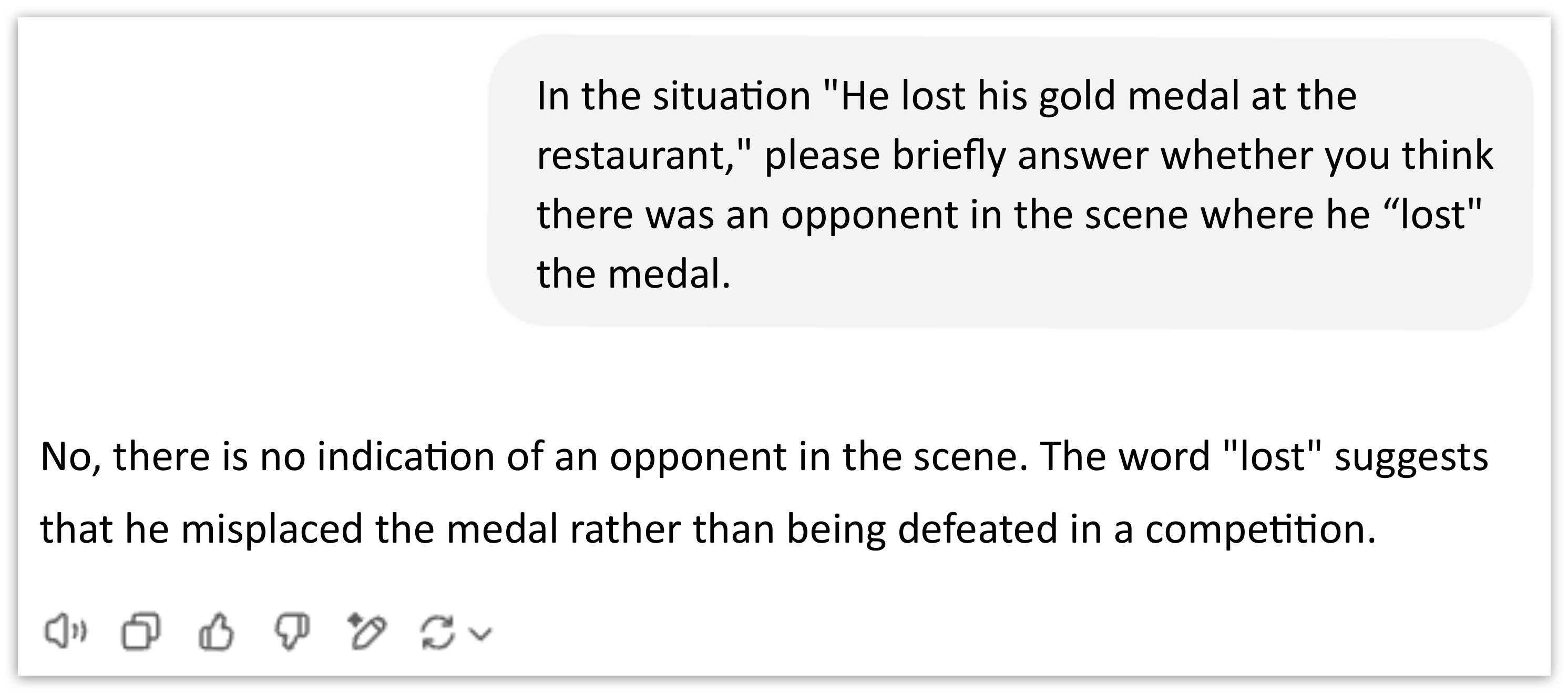}
        \\\vspace{1ex}
    \includegraphics[width=1.0\linewidth]{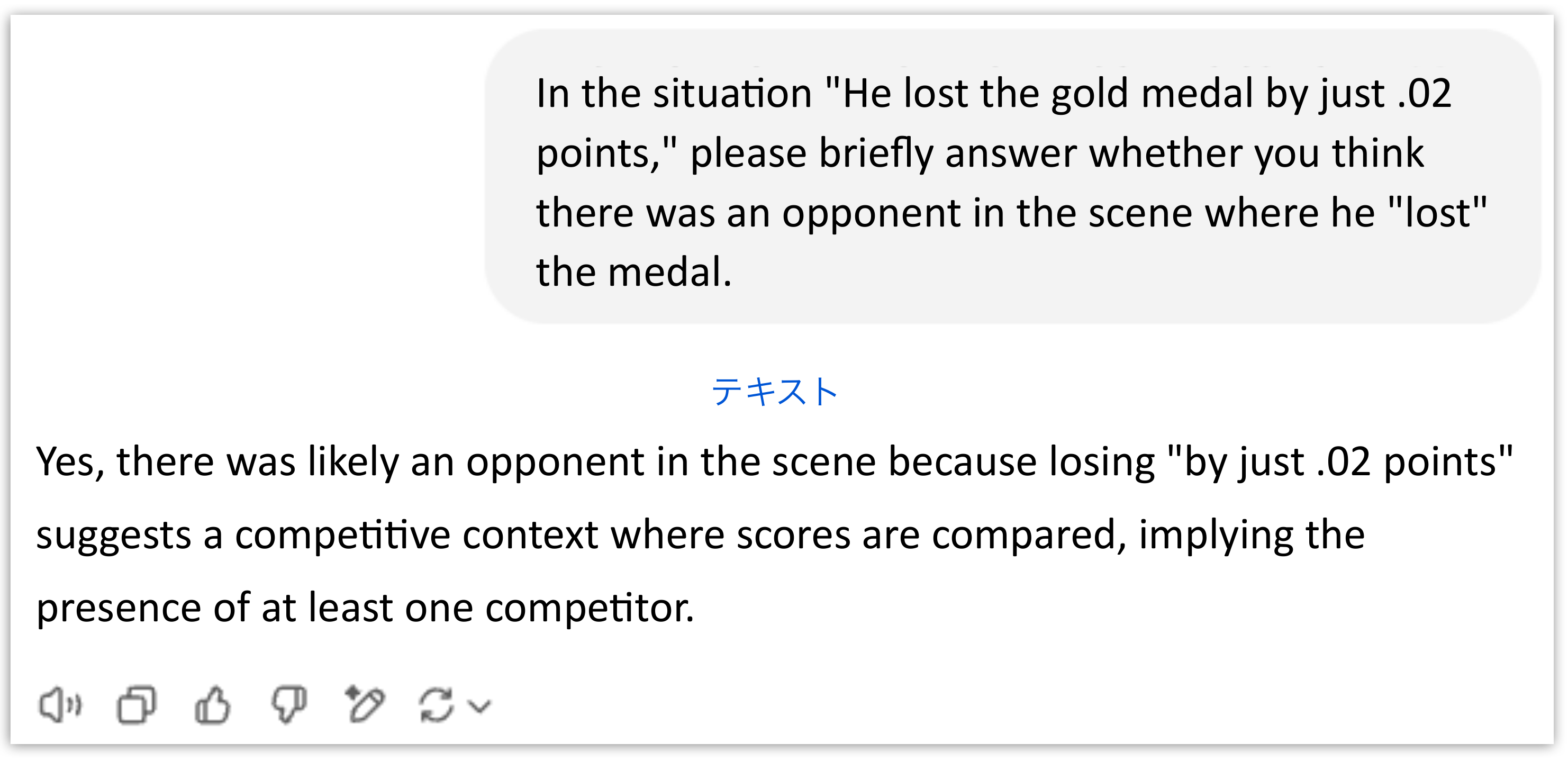}
        \caption{Examples of ChatGPT output showing the possibility of recognizing differences in semantic frames evoked by the same predicate \textit{lost}.}
    \label{fig:chat-gpt-demo}
    \vspace{-1.5ex}
\end{figure}

In contrast, large-scale causal language models (CLMs) such as the GPT and Llama series have not been used for semantic frame induction, despite their success in a wide range of language comprehension tasks~\cite{Llama2,GPT4}.
In fact, modern CLMs such as GPT-4o are potentially likely to understand semantic frames.
For example, as shown in Figure~\ref{fig:chat-gpt-demo},when ChatGPT receives the following prompt template with \texttt{[sentence]} replaced by Sentence (\ref{EX::Finish_competition}) and Sentence (\ref{EX::Losing}), respectively, it tends to answer ``Yes'' to the former and ``No'' to the latter.\footnote{We accessed \url{https://chatgpt.com/} on 2025-02-11 and used the GPT-4o model.}

\begin{description}
    \item[ \ Prompt template:]\ \\
    In the situation "\texttt{[sentence]}," please briefly answer whether you think there was an opponent in the scene where he "lost" the medal.
\end{description}

\noindent
This indicates that GPT-4o correctly understands that an opponent is present only in the situation represented by Sentence (\ref{EX::Finish_competition}).
It is thus conceivable that frame induction could be performed with higher accuracy by leveraging such CLMs.

\begin{figure*}[t]
    \centering
        \includegraphics[width=0.85\linewidth]{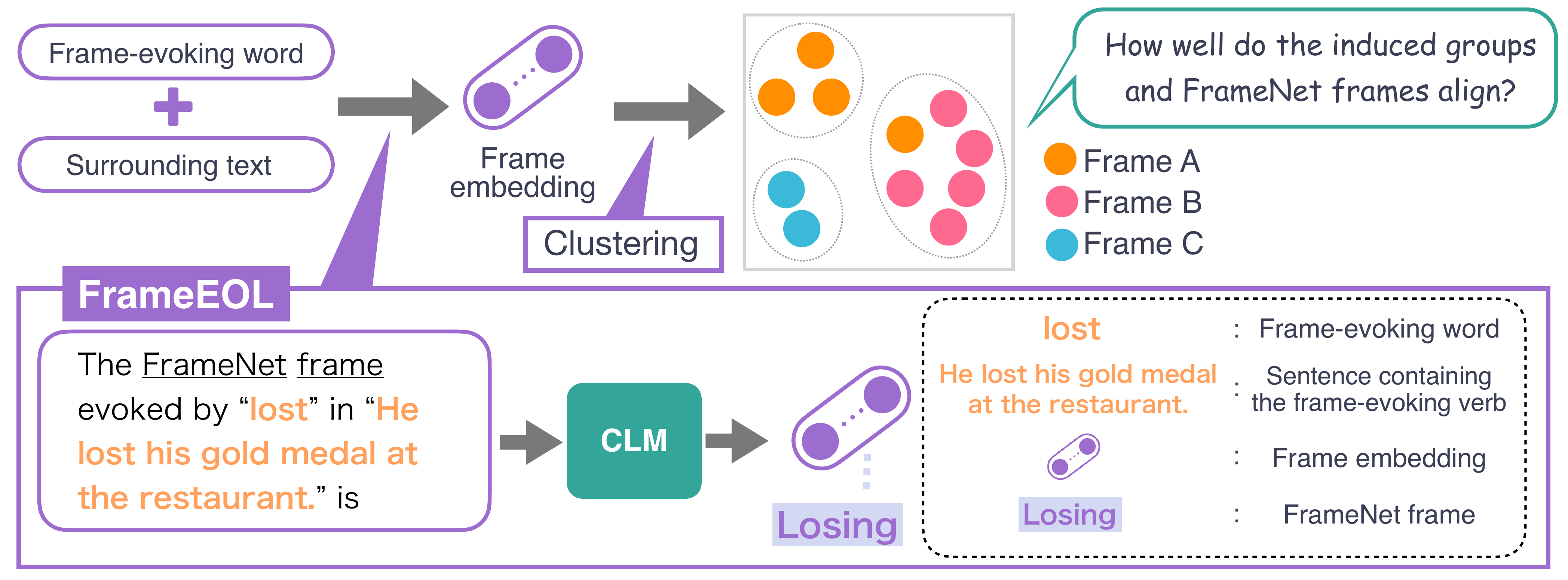}
        \caption{
    Overview of the frame induction process using FrameEOL and its evaluation.
    }
    \label{fig:decoder_emb}
\end{figure*}

Hence, in this paper, we propose a method for inducing semantic frames based on CLMs.
We first introduce \textbf{FrameEOL}, a prompt-based method for obtaining \textbf{F}rame \textbf{E}mbeddings that outputs \textbf{O}ne frame-name as a \textbf{L}abel representing the given situation.
\comment{FrameEOL extends PromptEOL \cite{prompteol}, a prompt-based sentence embedding method with an explicit one-word limitation, for semantic frame induction.}
Figure~\ref{fig:decoder_emb} shows an overview of FrameEOL and the overall frame induction process.
Given a frame-evoking word and its context, FrameEOL inputs a prompt to CLMs such that the next word will be the corresponding frame name in FrameNet.
As a result, the last token's embedding is considered to be the embedding that would represent that frame.
Furthermore, we apply in-context learning (ICL) and deep metric learning (DML) to obtain embeddings that are more suitable for semantic frame induction. 
After obtaining frame embeddings for the frame-evoking words in a given corpus, we cluster these embeddings to induce semantic frames.

\section{Related Work}
\subsection{Semantic Frame Induction}

\crcomment{
Semantic frame induction is a task that groups frame-evoking words, typically verbs, along with their contexts according to the semantic frames they evoke.
}

\crcomment{
Unlike semantic frame identification, which depends on a predefined set of frame labels, frame induction does not require such a predefined label set. 
This can be advantageous depending on the setting or application. 
First, when new concepts emerge, for example streaming in the sense of watching online videos, frame induction makes it possible to construct frames corresponding to those concepts ~\cite{tsujimoto-etal-2025-semantic}.
Second, in specific domains such as medicine or cooking, domain-specific concepts may exist and the required granularity of frames may differ. 
In such cases, it becomes necessary to perform frame induction that does not rely on a predefined label set.
}

\comment{
In recent years, methods leveraging contextualized embeddings have been proposed and shown to achieve strong performance~\cite{semeval2019-2_1st,semeval_BERT,semeval_ELMo,YAMADA_mask_2,YAMADA_23,mosolova-etal-2024}.}
They leverage MLMs such as BERT \cite{BERT} to obtain contextualized word embeddings. 
The methods can be broadly categorized into two groups: unsupervised and supervised approaches.
The unsupervised setting assumes that manually annotated frame information is unavailable, whereas the supervised setting assumes that manually annotated information is available for some frames.

In the unsupervised setting, several approaches have been proposed in recent years to group frame-evoking words via contextualized word embeddings derived from pre-trained language models \cite{semeval2019-2_1st,semeval_BERT,semeval_ELMo}.
Among these, \citet{YAMADA_mask_2} utilized two types of embeddings from BERT: the contextualized word embeddings of frame-evoking words; and the embeddings when the frame-evoking words are masked with \texttt{[MASK]} tokens. 
Furthermore, to address the tendency of existing methods to overly disperse examples across too many clusters, they introduced a two-step clustering approach.
While the commonly used one-step clustering approach clusters all examples at once, this approach first clusters examples by verb and subsequently clusters them across verbs, effectively limiting the number of clusters that examples of a single verb can belong to.
In this study, we will conduct experiments with both the one-step clustering approach and the two-step clustering approach in semantic frame induction.

In the supervised setting, \citet{YAMADA_23} demonstrated that semantic frame induction can be performed with even higher performance by applying deep metric learning (DML)~\cite{kaya-2019-deep}.
This approach addresses the limitation that embedding spaces obtained solely from pre-training do not always reflect meanings that align with human intuition about semantic frames. 
Through DML, the model is fine-tuned to place embeddings of verbs that evoke the same frame close together, while positioning embeddings of verbs evoking different frames farther apart in the semantic space. 
They employed several types of loss functions for DML including triplet loss, which optimizes triplets of anchor, positive, and negative examples to ensure the distance between the anchor and negative example is at least a margin larger than the distance between the anchor and positive example.

\subsection{Prompt-Based Text Embeddings}
\comment{In this study, we adapt prompt-based embedding techniques originally developed for sentence representations, to derive dense embeddings that represent semantic frames.
Therefore, in this section, we first provide an overview of existing methods for obtaining sentence embeddings.}

Traditional methods typically construct sentence embeddings with various strategies, including extracting representations from special tokens (e.g., \texttt{[CLS]}) or computing the average of all token embeddings~\cite{reimers-gurevych-2019-sentence,defsent,gao-etal-2021-simcse}.

More recently, prompt-based approaches have been explored to enhance the embedding quality~\cite{promptBERT,prompteol,metaEOL}.
\comment{PromptBERT~\cite{promptBERT} leverages a masked token prediction capability to obtain more effective sentence embeddings by using prompts such as ``This sentence: \texttt{[sentence]} means \texttt{[MASK]}'', which consolidate the sentence's semantic content into the \texttt{[MASK]} token representation.}
By leveraging masked token prediction, this approach dynamically contextualizes sentence meaning and has been shown to improve embedding quality.

\comment{
PromptEOL~\cite{prompteol} is a method to obtain sentence embeddings by utilizing the next token prediction capability of CLMs, and it demonstrates high performance with optimization via ICL or contrastive learning.
}
In this approach, a prompt is fed to a CLM to predict a single-word output that represents a given sentence's meaning.
The last token's final layer representation is used as the sentence embedding.
The following is an example of the prompt used in PromptEOL.
\begin{description}
\item[ \ Prompt template in PromptEOL:]\ \\ 
This sentence: "\texttt{[sentence]}" means in one word: "
\end{description}
The placeholder \texttt{[sentence]} represents the sentence to be embedded.

Both approaches design the input format to closely resemble the pre-training task, thereby enabling the model to better understand the task and obtain more effective embeddings. 
Inspired by these methods, we design tailored prompts for CLMs to effectively obtain frame embeddings.

\section{Proposed Method}

We induce semantic frames by obtaining the frame embeddings of frame-evoking words with FrameEOL and then clustering the frame-evoking words in a corpus based on the obtained frame embeddings.
Following \citet{YAMADA_mask_2}, we use two clustering methods: one-step and two-step clustering.
In this section, we first explain FrameEOL, a method for obtaining frame embeddings via a CLM; and then we describe how to optimize frame embeddings via  ICL and DML.

\subsection{FrameEOL}
\label{subsec:frame_emb_method}

Inspired by PromptEOL~\cite{prompteol}, we obtain the embedding of a frame-evoking word in a sentence by using a prompt tailored for frame induction.
In PromptEOL, the prompt is designed to predict a single word representing the given sentence as the next token, whereas in FrameEOL, the prompt is designed to predict a FrameNet frame name as the next token.
Specifically, FrameEOL uses the following prompt.
\begin{description}
\item[ \ Prompt template in FrameEOL:]\ \\ 
The FrameNet frame evoked by "\texttt{[verb]}" in "\texttt{[sentence]}" is
\end{description}
Here, \texttt{[verb]} denotes a placeholder for the frame-evoking verb for which we obtain the embedding, and \texttt{[sentence]} represents a placeholder for a sentence containing that frame-evoking verb.
We use the embedding from the final layer corresponding to the last token ``is'' as the frame embedding.

Note that FrameEOL assumes that pre-trained CLMs have learned knowledge of FrameNet, as can be seen from the inclusion of the word ``FrameNet'' in the above prompt.
This could be seen as some sort of leakage.
\comment{In Section~\ref{subsec:effect_framenet}, we remove the explicit term ``FrameNet'' from the input prompt and show that, under both ICL and DML settings, omitting this term has only a limited effect on performance. 
Since this study focuses on frame induction in a supervised setting,
we believe this potential leakage is not a major issue.}

\begin{figure}
\centering
\includegraphics[width=\linewidth]{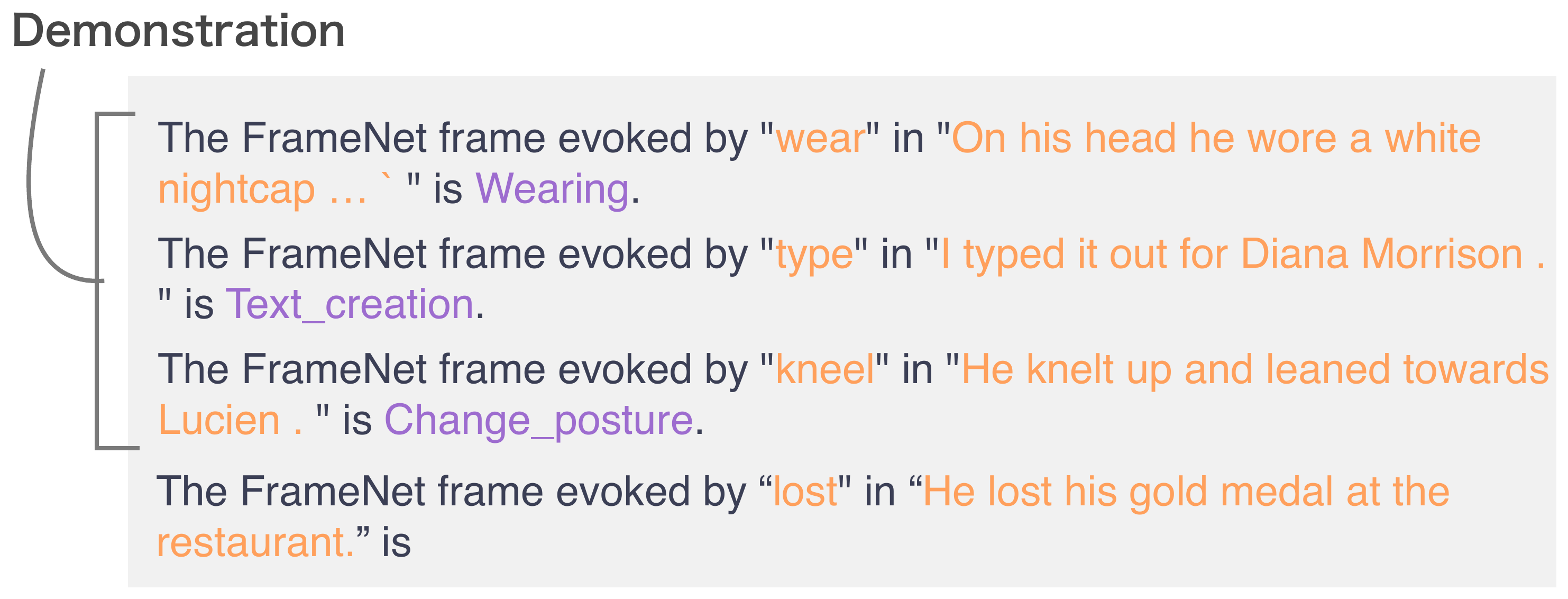}
\caption{Example of input in 3-shot ICL setting.}
\label{fig:few-shot}
\vspace{-1.5ex}
\end{figure}

\subsection{Frame Embedding Optimization via ICL}

Semantic frame induction has primarily been studied in English.
Its application to other languages remains challenging because of the limited availability of semantic frame resources.
ICL is a method that adapts CLMs to downstream tasks by incorporating a few demonstrations into the input~\cite{GPT3-fewshot}.
It has gained significant attention, particularly in scenarios with limited training data.
To achieve high-performance frame induction even for low-resource languages, we introduce a method to obtain frame embeddings based on ICL.
Figure~\ref{fig:few-shot} shows an example input provided to CLMs in a three-shot setting in our approach.
We first prepare examples comprising a frame-evoking word, its surrounding sentence, and the evoked frame name.
Next, we construct demonstrations by filling the FrameEOL prompt template: we fill the frame-evoking word into the \texttt{[verb]}, fill the surrounding sentence into the \texttt{[sentence]}, and then append the evoked frame name immediately afterward.
Following these demonstrations, we append an inference prompt that is constructed by filling the FrameEOL template with the target instance's frame-evoking word and its surrounding sentence without appending an evoked frame name.
By including such demonstrations, we expect to output embeddings that represent frame names, thus yielding distinct embeddings for each evoked frame.

\subsection{Frame Embedding Optimization via DML}

Previous studies on semantic frame induction using MLMs have demonstrated that, when supervised data are available, fine-tuning through DML is more effective than using pre-trained models as they are~\cite{YAMADA_23}. 
Given its proven effectiveness in prior work, we expect DML to be similarly effective here.
Hence, we use DML to enhance semantic frame induction using CLMs.

\begin{figure}[t]
    \centering
    \includegraphics[width=1.0\linewidth]{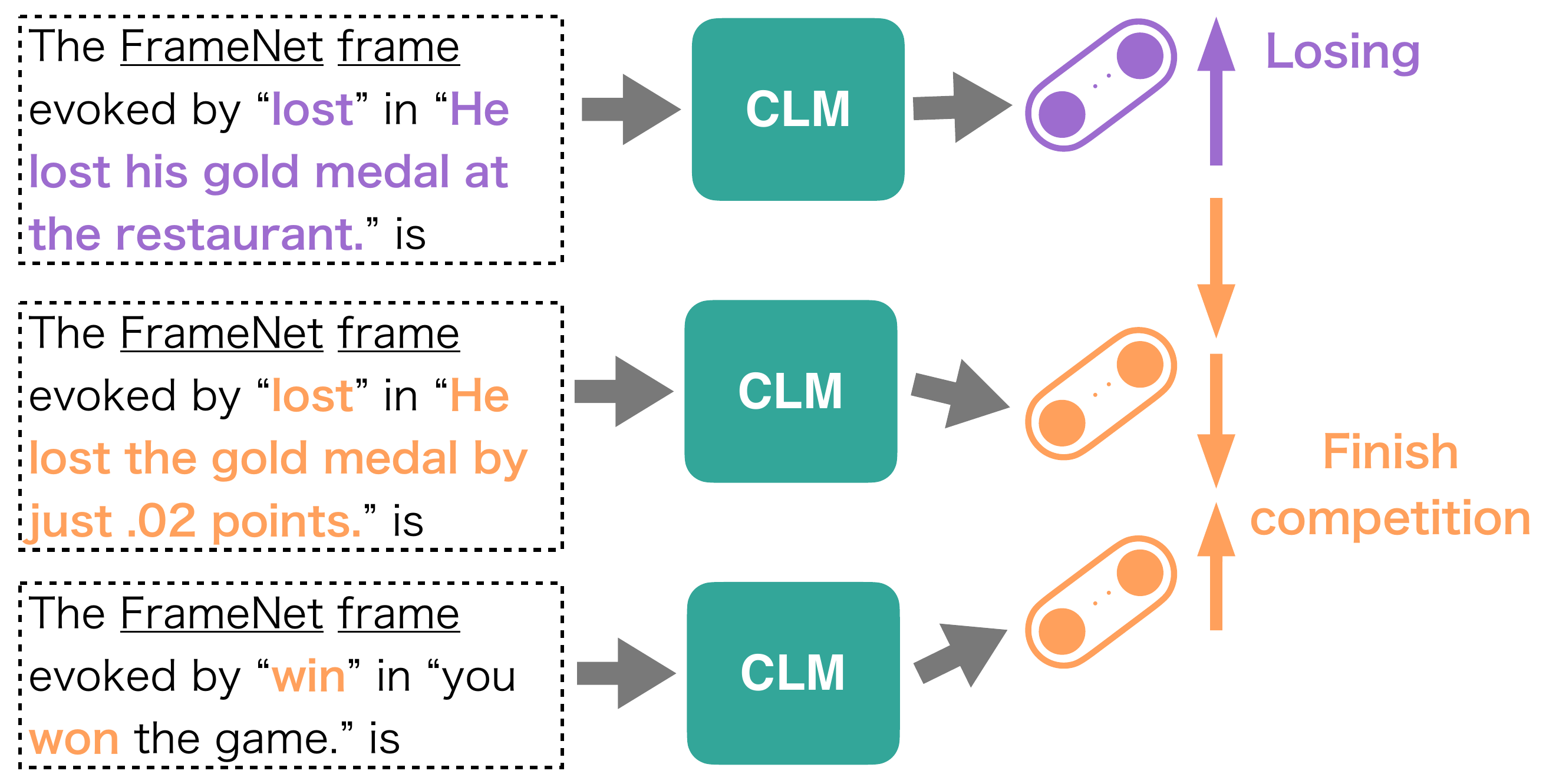}
    \caption{
Overview of DML: embeddings of the same frame are trained to be closer, while embeddings of different frames are trained to be farther apart.
    }
    \label{fig:fml}
    \vspace{-1.5ex}
\end{figure}

Given its demonstrated effectiveness~\cite{YAMADA_23}, we use the triplet loss:
\begin{equation}
\mathcal{L}_\mathrm{tri} = max(D(x_{a}, x_{p})-D(x_{a},x_{n})+m,0),
\label{equ:enc_triplet}
\end{equation}
\comment{
where $x_a$, $x_p$, and $x_n$ are the frame embeddings of the anchor, positive, and negative instances, respectively. 
The margin $m$ is a tunable hyperparameter, and $D(\cdot,\cdot)$ denotes a distance metric. }
In this study, we use the Euclidean distance of normalized frame embeddings $x$, following \citet{YAMADA_23}. 
Figure \ref{fig:fml} shows an overview of our DML application.

Because we use large-scale CLMs, we adopt a parameter-efficient approach to fine-tune these models under a constraint of limited computational resources.
Specifically, we use Low-Rank Adaptation (LoRA), a method that makes only the introduced low-rank matrices in linear layers trainable~\cite{LoRA}.
LoRA has been shown to achieve performance comparable to that of full fine-tuning while significantly reducing the number of trainable parameters.

\section{Experiments on FrameNet}
We first conducted experiments on semantic frame induction using CLMs for English.
In these experiments, we used FrameNet, an English frame knowledge resource, as the dataset for evaluation.

\subsection{Evaluation Setup}
\label{sec:framenet_setup}

\paragraph{Dataset}

We used a dataset that we extracted from FrameNet 1.7~\cite{FrameNet} by selecting in which the frame-evoking word is a verb.
\crcomment{
For three-fold cross validation, the dataset was equally split into three subsets in a 1:1:1 ratio for training, development, and test. 
}
The splits ensured that all examples of the same verb were placed in the same subset, and that the proportion of polysemous verbs remained constant across all subsets.
The statistics for the overall dataset and each subset are provided in Appendix \ref{app:statis_data}.
The average number of instances across the three subsets was 27,537.
\crcomment{A key characteristic of the experimental setup is that the test set includes frames that were unseen during training.
For instance, in the FrameNet dataset, an average of 135.3 out of 434.3 frames per test set were unseen during training.}

\paragraph{\comment{Clustering Methods}}
In the semantic frame induction task, the instances of each verb are clustered by the semantic frame that it evokes. 
Following previous work~\cite{YAMADA_23,YAMADA_mask_2}, we employ two clustering methods in our experiments: one-step clustering and two-step clustering.
For one-step clustering, we used group-average clustering, whereas for two-step clustering, we used X-means~\cite{X-means} for the first step and group-average clustering for the second step. 
\comment{Details are shown in Appendix \ref{app:clu_setting}.}

\paragraph{Evaluation Methods}
We evaluated the semantic frame induction methods by comparing automatically created clusters of instances of verbs with gold clusters of instances of verbs evoking the same frames. 
For evaluation metrics, we used B-cubed Precision (\textsc{BcP}), Recall (\textsc{BcR}), and their harmonic mean, F-score (\textsc{BcF}) \cite{B-cubed}, which are representative metrics for this clustering task.
\comment{In our experiments, we use \textsc{BcF} as the primary evaluation metric.}

\begin{table*}[h!]
\small
\centering
\vspace{0.5ex}
\begin{tabular}{cccc lll lll} \bhline
\tabH\multirow{2}{*}{Model} & \multirow{2}{*}{Model Size} &\multicolumn{2}{c}{Training}& \multicolumn{3}{c}{One-Step Clustering} & \multicolumn{3}{c}{Two-Step Clustering} \\
& & Method & Data Size & \mini{BcP} & \mini{BcR} & BcF & \mini{BcP} & \mini{BcR} & BcF \\ \bhline
\multicolumn{10}{c}{MLM-based methods} \\ 
\hline
\tabH \multirow{2}{*}{BERT$_\mathrm{base}$} & \multirow{2}{*}{0.1B}
& - &-&\mini{41.1}&\mini{44.7}&42.8 & \mini{49.7} & \mini{65.8} & 56.5 \\ 
&& DML& 27,536 &\mini{61.8} & \mini{68.7} & 65.0 & \mini{65.6} &\mini{67.8} & 66.4\\ \hdashline
\multirow{2}{*}{BERT$_\mathrm{large}$} & \multirow{2}{*}{0.3B}
& - &-&\mini{32.8} & \mini{45.9} & 38.3 & \mini{51.2} & \mini{65.4} & 57.3 \\ 
&& DML &27,536&\mini{63.6} & \mini{69.0} & 66.2 & \mini{67.3} & \mini{69.2} & 67.8\\ \hdashline
\tabH \multirow{2}{*}{ModernBERT$_\mathrm{base}$} & \multirow{2}{*}{0.1B}
& - &-&\mini{19.6}&\mini{42.8}& 27.0 & \mini{50.7} & \mini{61.4} &55.4  \\ 
&& DML& 27,536 &\mini{59.7} & \mini{69.7} & 64.3 & \mini{65.1} &\mini{68.4} & 66.5 \\ \hdashline
\multirow{2}{*}{ModernBERT$_\mathrm{large}$} & \multirow{2}{*}{0.3B}
& - &-&\mini{29.4}&\mini{45.1}& 35.6 & \mini{51.6} &\mini{62.1} & 56.2\\
&& DML& 27,536&\mini{63.5}&\mini{73.1}&\textbf{67.9} & \mini{68.2} & \mini{68.7} & 68.2 \\  \hdashline
\multirow{2}{*}{RoBERTa$_\mathrm{large}$} & \multirow{2}{*}{0.3B}
& - &-&\mini{21.9} & \mini{58.0} & 31.7 & \mini{52.9} & \mini{63.8} & 57.6 \\
&& DML&27,536&\mini{62.9} & \mini{73.7} & \textbf{67.9} & \mini{70.4} & \mini{69.2} & \textbf{69.6} \\
\bhline
\multicolumn{10}{c}{CLM-based methods (ours)} \\ 
\hline
\tabH \multirow{5}{*}{Gemma 3} & \multirow{5}{*}{12B}
& \multirow{4}{*}{ICL} &0 & \mini{9.1} & \mini{66.7} & 16.0 & \mini{53.3} & \mini{39.6} & 45.4\\ 
&& &5 & \mini{62.2} & \mini{55.4} & 58.6 & \mini{64.0} & \mini{53.8} & 58.4\\ 
& & &10 & \mini{63.0} & \mini{54.9} & 58.7 & \mini{64.5} & \mini{53.7} & 58.6 \\ 
& & &20 & \mini{63.0} & \mini{56.7} & 59.7 & \mini{64.8} & \mini{54.5} & 59.2\\  \cdashline{3-10}
\tabH && DML &27,536&\mini{\textbf{69.3}}&\mini{74.7}&\textbf{71.9}&\mini{72.2}&\mini{\textbf{69.4}}&70.6\\ 
\hdashline
\tabH \multirow{5}{*}{Llama 3.1} & \multirow{5}{*}{8B}
&  \multirow{4}{*}{ICL} &0& \mini{20.8}&\mini{45.1}&28.4&\mini{43.4}&\mini{48.8}&46.0 \\ 
&&  &5&\mini{55.9}&\mini{58.3}&57.0&\mini{49.0}&\mini{62.8}&54.9 \\ 
& &  &10&\mini{56.6}&\mini{59.0}&57.7&\mini{50.9}&\mini{62.9}&56.1\\ 
& & &20 & \mini{56.8}&\mini{60.1}&58.3&\mini{51.5}&\mini{63.3}&56.7\\ 
\cdashline{3-10}
\tabH  &&DML  &27,536&\mini{67.2}&\mini{\textbf{74.9}}&70.8&\mini{\textbf{74.2}}&\mini{68.3}&\textbf{70.9} \\ 
\bhline
\end{tabular}
\caption{
The experimental results of semantic frame induction on FrameNet. 
The scores were averaged over a three-fold cross-validation, and then multiplied by 100.
}
\label{tab:result_dec-dml}
\renewcommand{\arraystretch}{1.0}
\vspace{-1.5ex}
\end{table*}

\paragraph{Training Settings}
In our experiments, we used widely recognized, general-purpose pre-trained language models from the Gemma and Llama series. 

Specifically, we conducted experiments with Gemma 3-12B\footnote{\url{https://huggingface.co/google/gemma-3-12b-it}}~\cite{gemma_2025}, 
and Llama 3.1-8B\footnote{\url{https://huggingface.co/meta-llama/Llama-3.1-8B}}~\cite{llama3}.
\comment{Experiments on smaller models are presented in Appendix \ref{app:smaller_exp}, due to space constraints.}
We conducted fine-tuning with LoRA \comment{by attaching LoRA adapters to all linear layers within the model.}
We set LoRA rank $r=8$ and $\alpha=32$, and training for 20 epochs via the AdamW optimizer~\cite{loshchilov-2017-decoupled} with a batch size of 32.
We explored triplet loss margins of \{0.1, 0.2, 0.5, 1.0\} and learning rates of \{3e-5, 5e-5, 1e-4\}.
All hyperparameters were determined using the development set.
\comment{Training cost details are shown in Appendix~\ref{app:training_cost}.}

In the ICL experiments, the demonstrations included in the prompt were randomly sampled from the training data.
We experimented with 5, 10, and 20 demonstrations, conducting four runs with different random seeds to account for sampling variance.
We set the maximum sequence length to 2048 and truncated the beginnings of input sequences when necessary.\footnote{\comment{According to the Llama tokenizer, an average token length of 26.48 per example, and the prompt token length was 13}}
\comment{To prevent truncation, we replaced any demonstration exceeding 1,900 tokens with another randomly selected example.}

\paragraph{Comparison Methods}
\crcomment{
For comparison, we evaluated the method leveraging mask embeddings to obtain contextualized word embeddings~\cite{YAMADA_mask_2,YAMADA_23}, and also varied the choice of pre-trained models, including BERT$_\mathrm{base}$, BERT$_\mathrm{large}$, ModernBERT$_\mathrm{base}$, ModernBERT$_\mathrm{large}$, and RoBERTa$_\mathrm{large}$.
ModernBERT is a powerful MLM that incorporates optimization techniques recently applied to CLMs~\cite{modernbert}.
}
We evaluated methods without fine-tuning and methods with fine-tuning by DML using the triplet loss.

\subsection{Results}
Table~\ref{tab:result_dec-dml} summarizes the results.
\comment{
Our proposed FrameEOL-based method achieved the highest performance among all approaches by applying DML to the model.
Specifically, in the one-step clustering setting, Gemma 3 achieved a BcF score of 71.9 and Llama 3.1 scored 70.8---both substantially higher than the best encoder model's score of 67.9. 
In the two-step clustering setting, Gemma 3 and Llama 3.1 attained BcF scores of 70.6 and 70.9, respectively, again surpassing the MLM-based method's score of 69.6.}

The performance of the methods using ICL tended to improve slightly as the number of demonstrations increased.
The results confirmed that the CLM-based methods performed worse than the MLM-based methods when training data were not used, but the CLM-based methods achieved slightly better performance than the MLM-based methods when at least 20 instances were prepared. 
Specifically, the MLM-based method attained a score of 57.6 in the no-tuning setting, whereas the CLM-based method attained scores of 59.7 and 58.3 in the 20-shot ICL setting.
The ICL-based methods exhibit particular potential for use with low-resource language.

When using DML or ICL, no performance improvement was observed with two-step clustering.
This suggests that the learning process optimized the semantic embedding space sufficiently for semantic frames, thus enabling effective clustering of verb instances even with the one-step approach.
\subsection{Effect of ``FrameNet'' in the Prompt}
\label{subsec:effect_framenet}
FrameEOL explicitly incorporates FrameNet semantics by instructing the prompt to generate FrameNet frames for subsequent tokens in the prompt.
Given that the CLM's pre-training resources likely include FrameNet, the model's ability to perform reasonably well on few-shot semantic frame induction can be attributed to its potential use of FrameNet-related knowledge acquired during the pre-training phase. 
In this section, we examine the performance impact of the following input template, which omits the term ``FrameNet'' from the FrameEOL template.
\begin{description}
    \item[Prompt template without ``FrameNet'':]\ \\
The frame evoked by "\texttt{[verb]}" in "\texttt{[sentence]}" is
\end{description}
\comment{If the model can achieve comparable performance without the explicit term ``FrameNet'', then we can rule out the possibility that it is solving the task by leaking FrameNet information acquired during pre-training.}

\begin{table}[]
\setlength{\tabcolsep}{0.7ex}
 \small
    \centering
    \begin{tabular}{llc ll}
    \bhline
        \tabH \multirow{2}{*}{Model} & Training & ``FrameNet''   & One-Step  & Two-Step  \\        
        &Setting& in input  & BcF & BcF \\
        \bhline
        \tabH  \multirow{6}{*}{\shortstack{Gemma 3\\-12B}} & \multirow{2}{*}{-}&Yes & \textbf{16.0}    & \textbf{45.4}\\
        &&No& 13.5 \tiny{(‒1.5)}& 43.9 \tiny{(‒1.5)} \\
                \cline{2-5}
        \tabH & \multirow{2}{*}{\shortstack{ICL\\\mini{20-shot}}}&Yes &\textbf{59.7} & 59.2\\
        & &No& 59.6 \tiny{(‒0.1)} & \textbf{59.5} \tiny{(+0.3)} \\
                \cline{2-5}
        \tabH  & \multirow{2}{*}{DML}  &Yes & \textbf{71.9} & 70.6\\
        && No & 71.2 \tiny{(‒0.7)} & \textbf{70.8} \tiny{(+0.2)} \\
                \hline
                                                                                \tabH  \multirow{6}{*}{\shortstack{Llama 3\\-8B}} & \multirow{2}{*}{-}&Yes &\textbf{28.4} & \textbf{46.0}\\
        &&No&8.8 \tiny{(‒19.6)}& 35.9 \tiny{(‒10.1)}\\ \cline{2-5}
                \tabH   & \multirow{2}{*}{\shortstack{ICL\\ \mini{20-shot}}} &Yes &\textbf{58.3} & \textbf{56.7} \\
        & &No&58.2 \tiny{(‒0.1)} &55.8 \tiny{(‒0.9)}\\
                \cline{2-5}
        \tabH  & \multirow{2}{*}{DML}  &Yes & \textbf{70.8} & \textbf{70.9} \\
        &&No&70.5 \tiny{(‒0.3)}&  70.4 \tiny{(‒0.5)} \\
                \bhline
    \end{tabular}
    \caption{Differences in performance when the term ``FrameNet''was excluded from the input. 
        }
    \label{tab:no-framename}
    \vspace{-1.5ex}
\end{table}
Table~\ref{tab:no-framename} lists the results. 
\comment{
In the setting without fine‐tuning, the effect on performance of removing the term ``FrameNet'' from the input prompt varies by model; however, under both the ICL and DML settings, this effect remains limited.
These findings demonstrate that FrameNet information is not directly leaked from the pre-training corpus through the use of the term ``FrameNet''.
In other words, even if the pre-training corpus contains no explicit FrameNet annotations or related content, a CLM can still achieve high performance in frame induction once it is provided with sufficient annotated training data.
}

\section{Experiments on Japanese FrameNet}
For languages besides English, the limited scale of frame knowledge resources has constrained the development of semantic frame induction methods.
Consequently, most research in this field has been conducted in English. For instance, Japanese FrameNet~\cite{JFN}, a frame knowledge resource for Japanese, contains only 3,130 examples where verbs serve as frame-evoking words.
This is a very small dataset, representing only about 3.2\% of the number of similar examples in FrameNet.
The scarcity of training data makes it challenging to effectively train deep learning models, which has hindered the advancement of semantic frame induction methods in non-English languages.

In this study, to explore the performance of semantic frame induction for non-English languages, we also performed experiments using Japanese FrameNet for training and evaluation.
Large-scale CLMs have demonstrated strong multilingual capabilities, even when they are primarily English-centric~\cite{GPT3-fewshot,multiq}.
Accordingly, we investigated whether leveraging these multilingual capabilities can be beneficial for semantic frame induction, even when the amount of training data is limited.

\begin{table*}[h!]
 \small
\centering

\begin{tabular}{cccc lll lll}
\bhline
\tabH \multirow{2}{*}{Model}&\multirow{2}{*}{Model Size} &\multicolumn{2}{c}{Training}&\multicolumn{3}{c}{One-Step Clustering} & \multicolumn{3}{c}{Two-Step Clustering} \\
&&Method&Data Size& \mini{BcP}& \mini{BcR}& BcF& \mini{BcP} & \mini{BcR}& BcF \\
\bhline
\multicolumn{10}{c}{MLM-based methods} \\
\hline
\tabH \multirow{2}{*}{Japanese BERT$_\mathrm{base}$} & \multirow{2}{*}{0.1B} & - & - &\mini{42.8}& \mini{47.7}&44.6&\mini{47.3}&\mini{57.9}&51.8 \\
&& DML & 1,100 & \mini{55.1}& \mini{60.2} &57.1&\mini{54.3} &\mini{63.1} &58.2 \\ \hdashline
\tabH \multirow{2}{*}{\shortstack{Japanese \\ModernBERT$_\mathrm{base}$}} & \multirow{2}{*}{0.1B} & - & - &\mini{38.8}& \mini{42.7}&40.5&\mini{53.7}&\mini{54.5}&	54.0 \\
&& DML & 1,100 & \mini{59.0}& \mini{61.3} &\textbf{60.0}&\mini{51.6} &\mini{63.0} &	56.6 \\ \hdashline
\tabH \multirow{2}{*}{\shortstack{Japanese \\ModernBERT$_\mathrm{large}$}} & \multirow{2}{*}{0.3B} & - & - &\mini{36.8}& \mini{40.0}&38.2&\mini{50.6}&\mini{55.5}&	52.7 \\
&& DML & 1,100 & \mini{56.8}& \mini{63.5} &59.8&\mini{54.6} &\mini{63.1} &	\textbf{58.4} \\
\bhline
\multicolumn{10}{c}{CLM-based methods (ours)} \\
\hline
\tabH \multirow{5}{*}{Gemma 3}& \multirow{5}{*}{12B} & \multirow{4}{*}{ICL} & 0 & \mini{16.9} & \mini{21.9} & 18.7 & \mini{39.9} & \mini{45.0} & 42.3\\ 
&&  &5 & \mini{55.6} & \mini{61.9} & 58.5 & \mini{57.4} & \mini{60.9} & 59.0 \\ 
& &  &10&\mini{56.4} & \mini{61.0} & 58.5 & \mini{57.8} & \mini{61.1} & 59.4 \\ 
& &  &20&\mini{56.7} & \mini{60.2} & 58.3 &\mini{\textbf{57.9}} & \mini{60.8} & 59.3 \\ \cdashline{3-10}
\tabH && DML & 1,100 & \mini{51.9} &\mini{66.2} & 57.9 & \mini{53.8} & \mini{66.2} & 59.4 \\
\hdashline
\tabH \multirow{5}{*}{Llama 3.1}& \multirow{5}{*}{8B} & \multirow{4}{*}{ICL} & 0 & \mini{18.9} & \mini{43.7} & 26.2 & \mini{47.2} & \mini{50.2} & 48.5 \\
&&  &5& \mini{55.7} & \mini{65.1} & 59.9 & \mini{51.2} & \mini{65.5} & 57.4 \\
& &  &10& \mini{56.9} & \mini{64.9} & 60.6 & \mini{51.8} & \mini{65.6} & 57.8 \\
& &  &20&\mini{56.8} & \mini{65.5} & 60.9 & \mini{51.6} & \mini{65.8} & 57.8 \\
\cdashline{3-10}
\tabH && DML & 1,100 &\mini{\textbf{57.0}}&\mini{64.1}&60.3&\mini{56.3}&\mini{65.3}&\textbf{60.3}\\ 
\hdashline
\tabH \multirow{5}{*}{LLM-jp-3}& \multirow{5}{*}{13B} & \multirow{4}{*}{ICL} & 0 & \mini{19.7} & \mini{32.7} & 24.5 & \mini{43.1} & \mini{55.1} & 47.9\\
&&  &5& \mini{53.1} & \mini{65.0} & 58.4 & \mini{53.0} & \mini{64.6} & 58.2\\
&&  &10& \mini{54.2} & \mini{65.8} & 59.4 & \mini{51.9} & \mini{66.2} & 58.1\\
&&  &20& \mini{53.8} & \mini{66.3} & 59.3 & \mini{51.7} & \mini{67.1} & 58.4\\
\cdashline{3-10}
\tabH && DML & 1,100 & \mini{56.9} & \mini{\textbf{66.7}} & \textbf{61.3} & \mini{53.4} & \mini{\textbf{66.6}} & 59.2\\
\bhline

\end{tabular}
\caption{
The Experimental results of semantic frame induction on Japanese FrameNet. 
The scores were averaged over a three-fold cross-validation, and then multiplied by 100.
}
\label{tab:ja_result}

\renewcommand{\arraystretch}{1.0}
\vspace{-1.5ex}
\end{table*}
\subsection{Experimental Setup}

Our proposed method, FrameEOL, is adaptable to any language by simply modifying the prompt. To obtain frame embeddings in Japanese, we used the following prompt, which is a Japanese translation of the prompt described in Section~\ref{subsec:frame_emb_method}.
\begin{description}
    \item[Prompt template for Japanese:]\ \\ \begin{CJK}{UTF8}{ipxm}
"\texttt{[sentence]}" 内の "\texttt{[verb]}" が喚起する\\FrameNetフレームは
    \end{CJK}
    \\ \vspace{-2ex}
    \\
    (\textbf{\small English Translation:} The FrameNet frame evoked by "\texttt{[verb]}" in "\texttt{[sentence]}” is)
\end{description}

\comment{Experiments were conducted with two general-purpose CLMs, Gemma 3-12B-it~\cite{gemma_2025} and Llama 3.1-8B~\cite{llama3}, and one CLM intended for use in Japanese, LLM-jp-3-13B-Instruct3\footnote{\url{https://huggingface.co/llm-jp/llm-jp-3-13b-instruct3}}~\cite{llm-jp}.}
\crcomment{
 LLM-jp-3-13B-Instruct3 was trained on a total of 2.1T tokens, which included Japanese data, and underwent instruction tuning and direct preference optimization~\cite{dpo}.
}
\comment{The experiments on smaller models are presented in Appendix~\ref{app:smaller_exp}.}
The other experimental settings for clustering methods, DML, and ICL were the same as those described in Section~\ref{sec:framenet_setup}.
The datasets and comparison methods used for evaluation were as follows.

\paragraph{Dataset}
We extracted Japanese FrameNet examples in which verbs evoked specific frames.
For three-fold cross-validation, these examples were divided into training, development, and test sets, ensuring no overlap of frames or evoking words across subsets.
For frames with three or more evoking words, overlaps between subsets were allowed to prevent the number of evoking words per frame from becoming too small, given the dataset's limited size.
Nevertheless, as the split was performed at the verb level, no single verb appeared in multiple subsets. 
The statistics for the overall dataset and each subset are provided in Appendix \ref{app:statis_data}.
The average number of instances across the three subsets was 1,043.
\crcomment{
In the Japanese FrameNet dataset, an average of 110.7 out of 137.0 frames per test set were unseen during training.}

\paragraph{Comparison Methods}
For comparison methods, we also evaluated MLM-based approaches that work without fine-tuning~\cite{YAMADA_mask_2} and with fine-tuning~\cite{YAMADA_23} to a Japanese BERT model \crcomment{and Japanese ModernBERT model.}
\crcomment{ The BERT model was the one released by Tohoku University,\footnote{\url{https://huggingface.co/cl-tohoku/bert-base-japanese-v3}} which has 110M parameters and was trained on the Japanese Wikipedia and CC-100 datasets. 
This model is widely used in Japanese natural language processing tasks.
The ModernBERT model,\footnote{\url{https://huggingface.co/sbintuitions/modernbert-ja-130m}} was trained with 4.39T tokens of Japanese and English data~\cite{modernbert-ja}, which is more than double the amount used for LLM-jp-3-13B-Instruct3.
}

\subsection{Experimental Results}
Table~\ref{tab:ja_result} summarizes the experimental results.
\crcomment{
Our proposed CLM-based methods demonstrated superior performance compared to the MLM-based methods, regardless of the training setting.
The highest overall performance was achieved in the one-step clustering setting using LLMjp-3, reaching a BcF score of 61.3, surpassing the MLM-based method’s score of 60.0.
In the two-step clustering setting, Llama 3.1 achieved the highest BcF score of 60.3, exceeding the MLM-based method’s score of 58.4.
Across both clustering approaches, the ICL setting using five or more few-shot examples demonstrated performance that was comparable to the method of fine-tuning MLM with DML.
}
This was likely due to the small data size, as the training data for Japanese FrameNet comprises only 1,100 examples, approximately 4\% of the 27,536 examples available in English FrameNet.

When we use ICL, sufficient performance can be achieved with as few as 20 examples, which is significantly smaller compared to the number of examples required for DML with MLM.
Hence, even in cases where only a very limited number of examples is available, these results demonstrate the possibility of achieving highly accurate automatic construction of semantic frame knowledge by creating a small number of examples and leveraging CLMs combined with ICL.

\section{Conclusion}
In this study, we proposed a method using causal language models (CLMs) for semantic frame induction of verbs.
By leveraging the characteristics of CLMs, our proposed method uses a sequence in which the next word is the frame name as input to the CLMs, and it uses the final layer embedding of the last token as the frame embedding representation. 
In experiments on FrameNet, deep metric learning (DML) exceeded the performance of existing methods using masked language models (MLMs) with DML. 
In-context learning (ICL) showed lower performance than existing methods using MLMs trained on a large amount of supervised data with DML , but better performance than existing methods without DML. 
\crcomment{
In experiments on Japanese FrameNet, the ICL method demonstrated performance comparable to methods based on MLMs fine-tuned with DML. 
}
These results suggest that, when automatically constructing semantic frame knowledge for languages with no or very limited training data, a promising approach would be to create about 20 training examples and use a method combining CLMs with ICL.

\section*{Limitations}
Our proposed CLM-based semantic frame induction method demonstrated high performance. 
Nevertheless, several limitations remain. 
In particular, our experiments were limited to English FrameNet and Japanese FrameNet, leaving it unclear how well our approach would generalize to other languages or corpora. 
Additionally, the use of large CLMs entails significant computational requirements, which could restrict the practical deployment of our method in resource-constrained environments.

\crcomment{
\section*{Acknowledgments}
We would like to express our gratitude to Dr. Kyoko Ohara of Keio University for providing the Japanese FrameNet data used in this study. This research was supported by JST FOREST Program JPMJFR216N.}

\bibliography{main}
\appendix
\section{Statistics of the datasets}
\label{app:statis_data}

\begin{table}[h]
\small
    \centering
    \begin{tabular}{crrr} \bhline
           \tabH & \#Instances & \#Frames & \#Verbs\\ \bhline
        \tabH $\text{Split}_0$& 28,117 & 459 &830 \\
        \tabH $\text{Split}_1$& 28,314 & 429 & 831 \\ 
        \tabH $\text{Split}_2$& 26,179 & 415 & 831 \\ \hline
         \tabH Full & 82,610 & 642&2,492 \\ \bhline
    \end{tabular}
    \caption{Statistics of the datasets used in the FrameNet experiments.}
    \label{tab:statistics_on_en_datasets}
    \vspace{-1.5ex}
\end{table}

\begin{table}[h]
\small

    \centering
    \begin{tabular}{crrr} \bhline
       \tabH  & \#Instances & \#Frames & \#Verbs\\ \bhline
       \tabH  $\text{Split}_0$& 1,046 & 112 &209 \\
       \tabH  $\text{Split}_1$& 1,055 & 154 & 294 \\ 
       \tabH  $\text{Split}_2$& 1,029 & 145 & 263 \\ \hline
         \tabH Full & 3,130 & 344&766 \\ \bhline
    \end{tabular}
    \caption{Statistics of the datasets used in the Japanese FrameNet experiments.}
    \label{tab:statistics_on_ja_datasets}
    \vspace{-1.5ex}
\end{table}

The statistics of the datasets used in the experiments are presented. Table \ref{tab:statistics_on_en_datasets} shows the statistics for FrameNet, and Table \ref{tab:statistics_on_ja_datasets} shows the statistics for Japanese FrameNet.
\crcomment{
The average proportion of polysemous verbs in the test sets, which can evoke different frames depending on the context, was 30.3\% for FrameNet and 19.1\% for Japanese FrameNet.
}

\section{Clustering Setting}
\label{app:clu_setting}
Following \citet{YAMADA_mask_2}, we determine all clustering thresholds on the development set. 
\paragraph{One-step clustering} We employ group-average agglomerative clustering and iteratively merge clusters until their total number equals the number of frames in the development set. We then set the threshold to the minimum inter-cluster distance observed at this final merge.
\paragraph{Two-step clustering}
We begin by applying X-Means clustering to each group of examples sharing the same frame-evoking word, automatically inferring the optimal cluster count while capping the maximum clusters \(k_{\max}\) to the highest number of unique frames per verb in the development set. 
We then perform group-average agglomerative merging on the resulting clusters until the proportion of same-lemma examples assigned to the same cluster matches the proportion recorded on the development set. The minimum inter-cluster distance at this matching merge point serves as the second-stage threshold.

\section{Evaluation on Smaller Models}
\label{app:smaller_exp}
We report the frame induction performance of our proposed method when applied to smaller variants of the models used in Sections~4 and 5: Gemma 3-1B\footnote{{\url{https://huggingface.co/google/gemma-3-1b-it}}}, Gemma 3-4B\footnote{{\url{https://huggingface.co/google/gemma-3-4b-it}}}, Llama 3.2-1B\footnote{\url{https://huggingface.co/meta-llama/Llama-3.2-1B}}, and Llama 3.2-3B\footnote{\url{https://huggingface.co/meta-llama/Llama-3.2-3B}}. 
Table~\ref{tab:appendix_en} shows the results on FrameNet, and Table~\ref{tab:appendix_ja} shows the results on Japanese FrameNet.
\clearpage

\begin{table*}[h]
\small
\centering
\vspace{0.5ex}
\begin{tabular}{cccc lll lll} \bhline
\tabH \multirow{2}{*}{Model} & \multirow{2}{*}{Size}  &\multicolumn{2}{c}{Training} & \multicolumn{3}{c}{One-Step} & \multicolumn{3}{c}{Two-Step} \\
& & Method &Data Size & BcP & BcR & BcF & BcP & BcR & BcF \\ \bhline
\tabH \multirow{10}{*}{Gemma 3}&\multirow{5}{*}{1B}&
\multirow{4}{*}{ICL}&0& 3.0 & 88.8 & 5.8 & 28.6 & 28.2 & 28.3 \\ 
 &&&5& 38.3 & 53.1 & 44.3 & 48.5 & 55.6 & 51.7 \\ 
&&&10& 43.7 & 54.2 & 48.3 & 47.5 & 56.9 & 51.7\\ 
&&&20 & 45.2 & 54.1 & 49.1 & 48.7 & 57.8 & 52.8 \\
\cline{3-10}
\tabH &&DML&27,536&63.0&67.8&65.3&65.6&63.1&64.2\\ 
\cline{2-10}
\tabH &\multirow{5}{*}{4B}&
 \multirow{4}{*}{ICL}&0& 10.2 & 68.7 & 17.8 & 48.9 & 49.2 & 49.0 \\ 
 &&&5& 57.2 & 54.4 & 55.7 & 53.3 & 57.0 & 54.8 \\ 
 &&&10& 58.3 &	56.3 & 57.2 & 55.0 & 58.0 & 56.3\\ 
 &&&20& 58.1 &	59.3 & 58.6 &56.9 & 59.5 & 57.9\\ 
 \cline{3-10}
 \tabH& &DML&27,536&\textbf{67.7} & 68.8 & 68.2 & 69.2 & \textbf{67.3} & 68.1\\ 

\hline
\tabH \multirow{10}{*}{Llama 3.2}&\multirow{5}{*}{1B}&
\multirow{4}{*}{ICL}&0&3.0&84.4&5.9&46.0&30.2&36.4 \\
  &&&5&24.9&57.4&34.1&44.2&59.4&50.6 \\ 
  &&&10&30.2&59.0&39.7&44.4&59.9&50.9\\ 
  &&&20 & 34.6&59.8&43.7&44.6&60.8&51.4\\ 
  \cline{3-10}
\tabH  &&DML&27,536&64.7&70.9&67.6&70.3&65.2&67.4\\
\cline{2-10}
\tabH &\multirow{5}{*}{3B}&
\multirow{4}{*}{ICL}&0& 12.6&69.1&21.3&54.8&45.1&49.5 \\
 &&&5&44.3&59.4&50.6&47.7&64.2&54.7\\ 
 &&&10&47.3&59.4&52.5&48.4&62.9&54.6\\ 
 &&&20 & 48.8&60.3&53.9&48.2&63.3&54.7\\ 
 \cline{3-10}
 \tabH&&DML&27,536&66.1 & \textbf{73.2}&\textbf{69.4}&\textbf{74.1}&65.5&\textbf{69.2}\\ 
\bhline
\end{tabular}
\caption{
Frame induction performance of smaller models on FrameNet.
The scores were averaged over a three-fold cross-validation, and then multiplied by 100.
}
\vspace{-1.5ex}
\label{tab:appendix_en}
\end{table*}

\begin{table*}[h!]
 \small
\centering
\vspace{0.5ex}
\begin{tabular}{cccc lll lll} \bhline
\tabH \multirow{2}{*}{Model} & \multirow{2}{*}{Size}&\multicolumn{2}{c}{Training} & \multicolumn{3}{c}{One-Step} & \multicolumn{3}{c}{Two-Step} \\
& & Method &Data Size & BcP & BcR & BcF & BcP & BcR & BcF \\ \bhline
\tabH \multirow{10}{*}{Gemma 3}&\multirow{5}{*}{1B}&
\multirow{4}{*}{ICL}
&0& 16.8 & 19.2 & 17.8 & 37.9 & 44.0 & 40.7\\
&&&5&45.8 & 60.8 & 52.1 & 50.8 & 60.0 & 54.9\\
&&&10&46.8 & 61.8 & 53.1 & 49.1 & 61.7 & 54.6\\
&&&20&49.0 & 62.4 & 54.8 & 48.9 & 62.9 & 54.9\\
\cline{3-10}
\tabH &&DML&1,100&54.1 & 62.5 & 57.8 & 52.4 & 63.9 & 57.5\\ 
\cline{2-10}
\tabH &\multirow{5}{*}{4B}&
 \multirow{4}{*}{ICL}
 &0&16.5 & 20.4 & 17.9 & 40.6 & 47.2 & 43.6\\
 &&&5&52.4 & 65.5 & 58.2 & 55.1 & 63.3 & 58.8\\
 &&&10&52.5 & 66.0 & 58.4 & 56.2 & 63.5 & 59.5\\
 &&&20&53.6 & 66.6 & 59.2 & 55.5 & 64.0 & 59.4\\
 \cline{3-10}
 \tabH &&DML&1,100&55.4 & 64.5 & 59.5 & 52.5 & \textbf{65.6} & 58.3\\ 
\hline
\tabH \multirow{10}{*}{Llama 3.2}&\multirow{5}{*}{1B}&
 \multirow{4}{*}{ICL}
 &0&3.0 & 84.4 & 5.9 & 46.0 & 30.2 & 36.4\\
 &&&5&44.2 & 55.4 & 49.0 & 44.1 & 63.5 & 52.1\\
 &&&10&48.7 & 57.8 & 52.7 & 44.8 & 63.2 & 52.3\\
 &&&20&51.8 & 61.6 & 56.2 & 46.7 & 64.0 & 53.9\\
 \cline{3-10}
 \tabH &&DML&1,100&55.1 & 63.2 & 58.8 & 52.9 & 64.2 & 58.0\\ 
\cline{2-10}
\tabH &\multirow{5}{*}{3B}&
 \multirow{4}{*}{ICL}
 &0&19.8 & 39.7 & 26.2 & 44.2 & 54.5 & 48.5\\
 &&&5&53.1 & 64.2 & 58.1 & 51.0 & 64.8 & 57.0\\
 &&&10&52.7 & 63.5 & 57.6 & 50.5 & 64.5 & 56.6\\
 &&&20&53.1 & 63.7 & 57.9 & 51.5 & 64.2 & 57.0\\
 \cline{3-10}
 \tabH &&DML&1,100&\textbf{55.5} & \textbf{65.7} & \textbf{60.1} & \textbf{55.9} & 64.9 & \textbf{60.0}\\ 
\bhline
\end{tabular}
\caption{
Frame induction performance of smaller models on Japanese FrameNet.
The scores were averaged over a three-fold cross-validation, and then multiplied by 100.
}
\vspace{-1.5ex}
\label{tab:appendix_ja}
\end{table*}

\clearpage

\section{Training Cost}
\label{app:training_cost}

\begin{table}[H]
\small
\centering
\begin{tabular}{lrrr}
\bhline
\tabH \multirow{2}{*}{Model} &\multirow{2}{*}{Size} & \multicolumn{2}{c}{\#Params} \\
&&All & Trainable\\
\bhline
\tabH \multirow{3}{*}{Gemma 3}&1B& 1,006,408,832 &  6,522,880 \\
&4B& 3,895,164,416 & 14,901,248 \\
&12B& 11,798,769,408 & 32,735,232 \\ \hline
\tabH \multirow{2}{*}{Llama 3.2}&1B&1,241,450,496&5,636,096 \\
&3B&3,224,906,752&12,156,928 \\ \hline
\tabH Llama 3.1&8B&7,525,896,192&20,971,520 \\
\hline
\tabH LLM-jp-3&13B &13,229,347,840 &31,293,440  \\
\bhline
\end{tabular}
\caption{Total number of parameters and trainable parameters after adding the LoRA layers.}
\label{tab:appendix_params}
\end{table}

\begin{table}[H]
\small
  \centering
  \label{tab:training_times}
  \begin{tabular}{l lr r}
    \bhline
    \tabH Dataset & Model &Size & Training Time\\
    \bhline
    \tabH \multirow{6}{*}{FrameNet} 
      & \multirow{3}{*}{Gemma 3}&1B &  180 min \\
      & &4B &  540 min\\
      & &12B &  1200 min\\ \cline{2-4}
                  \tabH & \multirow{2}{*}{Llama 3.2}&1B & 150 min\\
      & &3B & 360 min\\ \cline{2-4}
      \tabH &Llama 3.1 &8B & 720 min\\
    \hline
    \tabH \multirow{7}{*}{\shortstack{Japanese\\FramaNet}} 
      & \multirow{3}{*}{Gemma 3}&1B &  12 min \\
      & &4B &  30 min\\
      & &12B &  80 min\\ \cline{2-4}
                \tabH & \multirow{2}{*}{Llama 3.2}&1B & 12 min\\
      & &3B & 30 min\\\cline{2-4}
      \tabH &Llama 3.1 &8B & 60 min\\ \cline{2-4}
    \tabH &LLM-jp-3 & 13B & 60 min \\
    \bhline
  \end{tabular}
  \caption{Training time required for the DML experiments.}
  \label{tab:appendix_training-time}
\end{table}

In deep metric learning, LoRA weights were applied to all linear layers.
Table \ref{tab:appendix_params} shows the total and trainable parameters for each model after adding the LoRA layers.

The training was conducted on a single A100 GPU with 80GB VRAM, and the time required is shown in Table \ref{tab:appendix_training-time}.
For English FrameNet, an average of 27,537 instances were used for training; with a batch size of 32 over 20 epochs, the average number of training steps was 17,210.
For Japanese FrameNet, an average of 1,043 instances were used for training; with a batch size of 32 over 20 epochs, the average number of training steps was 652.

\end{document}